\documentclass[11pt,a4paper, table,xcdraw]{article}
\usepackage[hyperref]{acl2021}
\usepackage{times}
\usepackage{latexsym}
\usepackage{xcolor, colortbl}
\usepackage{changepage}

\usepackage{array}

\usepackage{microtype}
\usepackage{graphicx}

\usepackage{multirow}

\usepackage{fancyhdr}
\usepackage{subcaption}
\usepackage[english]{babel}
\usepackage{tabularx}

\title{Evaluation of In-Person Counseling Strategies To Develop Physical Activity Chatbot for Women} 

\author{
Kai-Hui Liang \\ Columbia University \\ kaihui.liang@columbia.edu
\And
Patrick Lange \\ University of California, Davis   \\ pllange@ucdavis.edu 
\And
Yoo Jung Oh \\ University of California, Davis\\  yjeoh@ucdavis.edu
\AND
Jingwen Zhang \\ University of California, Davis\\ jwzzhang@ucdavis.edu 
\And
Yoshimi Fukuoka \\ University of California, \\ San Francisco\\ Yoshimi.Fukuoka@ucsf.edu  
\And
Zhou Yu \\ Columbia University   \\ zy2461@columbia.edu
}

\date{}

\aclfinalcopy

\begin{document}
\maketitle

\begin{abstract}
Artificial intelligence chatbots are the vanguard in technology-based intervention to change people's behavior. To develop intervention chatbots, the first step is to understand natural language conversation strategies in human conversation. This work introduces an intervention conversation dataset collected from a real-world physical activity intervention program for women. We designed comprehensive annotation schemes in four dimensions (domain, strategy, social exchange, and task-focused exchange) and annotated a subset of dialogs. We built a strategy classifier with context information to detect strategies from both trainers and participants based on the annotation. To understand how human intervention induces effective behavior changes, we analyzed the relationships between the intervention strategies and the participants' changes in the barrier and social support for physical activity. We also analyzed how participant's baseline weight correlates to the amount of occurrence of the corresponding strategy. This work lays the foundation for developing a personalized physical activity intervention bot. 
\footnote{The dataset and code are available at \url{https://github.com/KaihuiLiang/physical-activity-counseling}}

\end{abstract}

\section{Introduction} 

Physical inactivity is a leading risk factor for premature death from noncommunicable diseases such as heart disease, stroke, and type 2 diabetes \cite{american2013american, murphy2013national}. Despite the known benefits of physical activity (PA) in reducing morbidity and mortality \cite{samitz2011domains, wen2011minimum}, physical inactivity is common among Americans. About 80\% of American adults do not meet the guidelines for both aerobic and muscle-strengthening activities \cite{clarke2019early}. Common reasons women are more likely than men to not meeting physical activity guidelines include lack of motivation, lack of social support, lack of time in exercising, etc. Effective interventions that can help women overcome these barriers and engage in more regular activity are needed to reduce multiple health risks.

Physical activity intervention programs have evolved with emerging digital and communication technologies \cite{vandelanotte2016past, case2015accuracy, mateo2015mobile, zhang2016support, zhang2015efficacy, zhang2017sub}. Recently, effective technology-based interventions have been published. For example, a pilot randomized clinical trial (RCT) of a mobile app-based online group intervention for African American young women \cite{zhang2019mobile} showed the online tracking and social support increased objectively measured daily physical activity in comparison to a control condition where participants only used the Fitbit for self-monitoring. Another RCT tested the use of a mobile app in conjunction with brief in-person counseling and found the combination increased objectively measured physical activity over three months compared to a control condition in which participants only used accelerometers \cite{fukuoka2011mped, fukuoka2019short}. 

These interventions lack the capacity to tailor the intervention messages to accommodate different individuals' needs and circumstances and automate such personalized messages through mobile technologies. Artificial intelligence (AI)-based chatbots are the vanguard in technology-based interventions, and they can deliver intervention messages and tailor contents to meet individual needs through natural conversations with no spatial or time restraints. 

The first step to develop physical activity intervention chatbots is to learn natural language conversation strategies from human-human conversations in physical activity intervention domains. Specifically, it is vital to understand how participants' and trainers' conversation strategies influence the outcomes and how trainers could adapt to different physical activity statuses, socio-demographics, and conversation behaviors to achieve better results.

In this research, we aim to address this question by analyzing a real-world intervention conversation dataset collected as a part of an effective physical activity intervention program for women \cite{fukuoka2011mped, fukuoka2019short}. Unlike the commonly used role-play dialog datasets, our dataset consists of actual dialogs between research staff (trainer) and study participants. We developed a comprehensive annotation scheme based on how the original intervention was organized to extract both social and persuasive conversational strategies. Then we manually annotated a set of 17 conversations with 7,808 sentences. After achieving high inter-rater reliability levels, we developed a BERT-based classifier to detect the whole unannotated dataset's strategy. Lastly, we analyzed which and to what extent specific conversational strategies decrease physical activity barriers and increase social support among the intervention participants from the first visit (baseline) to the 3-month visit. 

The following research questions guide our analysis: 
\textbf{RQ1}: Does using more barrier strategies by trainers and participants in the intervention session decrease participants' physical activity-related barriers?
\textbf{RQ2}: Does using more support strategies by trainers and participants in the intervention session increase participants' physical activity-related social support?
\textbf{RQ3}: Do participants with a heavier weight at baseline use more weight strategies in the intervention session than participants with lighter weight? 




This work's main contribution is that we created a real-world human-human intervention dialog dataset that can be used to build physical activity promotion dialog systems. We also developed and designed a set of comprehensive four dimension annotation schemes that can be leveraged to behavior-change dialogs. Lastly, our analysis revealed how trainers' and participants' usage of conversational strategies influence the outcome and how a physical activity intervention chatbot could better adapt to participants' individual needs.




\section{Related Work}

Applying AI chatbots to lifestyle modification programs (e.g., physical activity and diet promotion) has great potential to provide cost-effective, sustainable, and broadly applicable solutions and is anticipated to benefit health across application domains \cite{laranjo2018conversational, zhang2020artificial}. Previous studies that developed and tested the efficacy of chatbot-delivered physical activity and diet interventions have demonstrated the potential of using chatbots as a practical solution to promote positive behavior changes \cite{casas2018food, kramer2020components, maher2020physical, piao2020use, stephens2019feasibility}. Among these studies, some have demonstrated how theory-driven intervention strategies combined with AI chatbot technologies can effectively yield behavioral changes \cite{kramer2020components, piao2020use, stephens2019feasibility}. It was also shown that a chatbot could provide richer intervention when combined with behavior monitoring technology, such as mobile or wearable tracking tools that enable real-time monitoring of user activity \cite{kramer2020components, kunzler2019exploring}.

Such developments in physical activity and diet change promotion chatbots have contributed to our understanding of the feasibility and effectiveness of chatbot-delivered interventions. To this extent, the existing studies using chatbots for interventions mainly focused on examining the effectiveness of chatbot-delivered strategies (e.g., intervention messages) on physical activity and diet outcomes. Although users' conversational inputs can be valuable to successful interventions, previous studies lacked discussion of how users' conversational inputs during the interventions, such as their reflections of behaviors and environments, may have affected the outcomes \cite{kocielnik2018reflection}. Hence, a quantitative analysis of user responses to the chatbot's messages is necessary to better grasp the bot and users' conversational patterns and how they lead to positive outcomes. 

In this study, we investigate the effects of barrier and support strategies used by the trainer and participants during a 3-month physical activity intervention program and on the intervention outcomes (i.e., changes in participants' physical activity-related barriers and social support). In addition, we explore whether participants' baseline weight (i.e., one's weight before the intervention) would influence the amount of weight-related strategies they mentioned in the conversations.
\section{Dataset} 

This paper used the data collected from the mobile phone-based physical activity education program (mPED) study in community-dwelling women aged 25 to 69. The study protocol was approved by the University of California, San Francisco, Committee on Human Research, and the mPED Data and Safety Monitoring Board. Detailed descriptions of the study design and outcomes have been previously published \cite{fukuoka2011mped, fukuoka2019short}. In brief, the mPED trial was an unblinded, parallel randomized clinical trial (RCT) conducted with three groups (control, regular, and plus groups). In this study, we used the data from the intervention groups (regular and plus groups) who received the identical physical activity intervention, consisting of brief in-person counseling sessions, an accelerometer, and the mPED trial app for the first three months. 

At the baseline visit, research staff collected participants' sociodemographic information (e.g., age, education, marital status, employment, and racial/ethnicity), assessed participants' weight, and administered the Barriers to Being Physically Active Quiz and the Social Support and Exercise Survey. The Barriers to Being Physically Active Quiz developed by the Centers for Disease Control and Prevention (CDC) \cite{sallis1987development} is a 21-item measure assessing the following barriers to physical activity: 1) lack of time, 2) social influence, 3) lack of energy, 4) lack of willpower, 5) fear of injury, 6) lack of skill, and 7) lack of resources (e.g., recreational facilities, exercise equipment). Each domain contains three items, with a total score range of 0 to 63, with higher scores indicating more barriers. Respondents rate the degree of activity interference on a 4-point scale, ranging from 0=``very unlikely'' to 3 = ``very likely.'' The Social Support and Exercise Survey was used to assess both friend and family social support related to physical activity during the past three months \cite{sallis1987development}. The measure consists of two subscales (friend and family support subscales).  Each subscale has 13 items with 5-point Likert scales (ranging from 1=``none'' to 5=``very often''). The ratings of all 13 items were summed for a subtotal score. Scores can range from 13 to 65, with higher scores indicating more support.

Women who met eligibility criteria (\ref{sec:eligibility}) and were randomized to the intervention groups received brief in-person physical activity counseling by trained research staff. All counseling sessions were digitally recorded. The average length of the counseling was 28.8 (SD 6.6) minutes. We randomly selected 107 sessions and had the audio recordings transcribed verbatim by a professional transcriptionist. 
On average, the trainers and participants spoke 213.91 and 209.63 turns respectively per session. The average sentence length and the average words per sentence from the trainers (397.07 sentences and 10.02 words/sentence) are longer than the participants' (277.67 sentences and 5.99 words/sentence). This is understandable as the trainers were supposed to deliver physical activity educational content during the counseling. 

After three months, the Barriers to Being Physically Active Quiz and the Social Support and Exercise Survey were administered again to assess the changes (from 3 months to baseline) in these measures. Among the 107 transcribed dialogs, two dialogs were dropped due to missing survey results, 17 dialogs  (7,808 sentences) were randomly picked for annotation, and the remaining 88 dialogs (63,288 sentences) were used for classifier pretraining and data analysis.

Since releasing the original interview data is not approved by our IRB and HIPPA, we created and released 44 simulated dialogs (772 sentences) based on the original interview data for our community to use. (More statistics are listed in Appendix \ref{sec:simulated-dialog}).
\section{Annotation Scheme}
\begin{table*}[t!]
\centering
\fontsize{9}{9}\selectfont
 \begin{adjustwidth}{}{}
\begin{tabular}{|p{7.6cm}|p{1cm}|p{1.3cm}|p{1.1cm}|p{1.2cm}|p{1.1cm}|} 
\hline
  \textbf{Utterance} &
  \textbf{Domain} &
  \textbf{Strategy 1} &
  \textbf{Strategy 2} &
  \textbf{Social \newline Exchange} &
  \textbf{Task-Focused} \\ \hline
\rowcolor[HTML]{F3F3F3} 
T: 
  \textit{So again your long-term goal, you'll reach ten thousand steps at week seven and to maintain it from there.} &
  Goal &
  Goal &
  None &
  None &
  Give-GenInfo \\ \hline
\cellcolor[HTML]{FFFFFF}P:  
  \cellcolor[HTML]{FFFFFF}\textit{Okay.} &
  Goal &
  None &
  None &
  Agree &
  None \\ \hline
\rowcolor[HTML]{F3F3F3} 
T:  
  \textit{So how confident do you feel that you can meet your long-term each week?} &
  Goal &
  Self-efficacy &
  Goal &
  None &
  Ask-Opinion \\ \hline
\cellcolor[HTML]{FFFFFF}P:  
  \cellcolor[HTML]{FFFFFF}\textit{I feel confident.} &
  Goal &
  Self-efficacy &
  None &
  None &
  Give-Opinion \\ \hline
\rowcolor[HTML]{F3F3F3} 
T:  
  \textit{Okay, great.} &
  Goal &
  None &
  None &
  Agree &
  None \\ \hline
\rowcolor[HTML]{F3F3F3} 
T:  
  \textit{So to break it down a little bit more for you, ten minutes brisk walking is gonna give you about a thousand to twelve hundred steps. }&
  Goal &
  Monitoring &
  None &
  None &
  Give-GenInfo \\ \hline
\cellcolor[HTML]{FFFFFF}P:  
  \cellcolor[HTML]{FFFFFF}\textit{Okay.} &
  Goal &
  None &
  None &
  Agree &
  None \\ \hline
\rowcolor[HTML]{F3F3F3} 
T:  
  \textit{So, think about brisk walking as a pace where you can still carry a conversation, but you can't sing. } &
  Goal &
  Monitoring &
  None &
  None &
  Give-GenInfo \\ \hline
\rowcolor[HTML]{F3F3F3} 
T:  
  \textit{And then make sure you walk for at least ten to fifteen minutes each time.} &
  Goal &
  Monitoring &
  None &
  None &
  Give-GenInfo \\ \hline
\rowcolor[HTML]{F3F3F3} 
T:  
  \textit{And the reason for that is that's going to give you the most health benefits of physical activity when you do it.} &
  Benefit &
  Monitoring &
  Benefit &
  None &
  Give-GenInfo \\ \hline
\cellcolor[HTML]{FFFFFF}P:  
  \cellcolor[HTML]{FFFFFF}\textit{Yeah.} &
  Benefit &
  None &
  None &
  Agree &
  None \\ \hline
\rowcolor[HTML]{F3F3F3} 
T:  
  \textit{And some of the health benefits of physical activity, regardless of your BMI, are decreased risk of breast and colon cancer, coronary heart disease, high blood pressure, diabetes, stress, depressive symptoms, osteoporosis.} &
  Benefit &
  Benefit &
  None &
  None &
  Give-GenInfo \\ \hline
\rowcolor[HTML]{F3F3F3} 
T:  
  \textit{And then increased energy level, emotional wellbeing, self-confidence, body image, and weight management, okay? }&
  Benefit &
  Benefit &
  None &
  None &
  Give-GenInfo \\ \hline
\cellcolor[HTML]{FFFFFF}P:  
  \cellcolor[HTML]{FFFFFF}\textit{Okay.} &
  Benefit &
  None &
  None &
  Agree &
  None \\ \hline
\rowcolor[HTML]{F3F3F3} 
T:  
  \textit{So which benefits of physical activity are the most important to you? }&
  Benefit &
  Benefit &
  None &
  None &
  Ask-PerInfo \\ \hline
\cellcolor[HTML]{FFFFFF}P:  
  \cellcolor[HTML]{FFFFFF}\textit{To me it's a decreased risk of breast and colon cancer.} &
  Benefit &
  Benefit &
  None &
  None &
  Give-PerInfo \\ \hline
\rowcolor[HTML]{F3F3F3} 
T:  
  \textit{Mm-hmm (affirmative), great. }&
  Benefit &
  None &
  None &
  Agree &
  None \\ \hline
\rowcolor[HTML]{F3F3F3} 
T:  
  \textit{All right, so a lot of women who have been inactive identify different barriers to physical activity, some of which are like lack of time, lack of social support, family obligations, maybe their neighborhood isn't great for walking.} &
  Barrier &
  Barrier &
  None &
  None &
  Give-GenInfo \\ \hline
\rowcolor[HTML]{F3F3F3} 
T:  
  \textit{Lack of resources, maybe they feel like they can only really workout in a gym, and they don't have the money.} &
  Barrier &
  Barrier &
  None &
  None &
  Give-GenInfo \\ \hline
\cellcolor[HTML]{FFFFFF}P:  
  \cellcolor[HTML]{FFFFFF}\textit{Yeah.} &
  Barrier &
  None &
  None &
  Agree &
  None \\ \hline
\rowcolor[HTML]{F3F3F3} 
T:  
  \textit{So tell me about some of the barriers that have been for you.} &
  Barrier &
  Barrier &
  None &
  None &
  Ask-PerInfo \\ \hline
\cellcolor[HTML]{FFFFFF}P:  
  \cellcolor[HTML]{FFFFFF}\textit{Lack of support, yeah, I used to have a couple of walking partners who are not there anymore.} &
  Barrier &
  Barrier &
  Support &
  None &
  Give-PerInfo \\ \hline
\end{tabular}
\caption{Example dialog snippet with the four dimension annotations. (T: trainer, P: participant)}
\label{tab:annotation-example}
 \end{adjustwidth}
\end{table*}

\begin{table*}[t!]
\centering
\fontsize{9}{9}\selectfont
\begin{tabularx}{\textwidth}{lp{6.3cm}X}
\hline
\textbf{Strategy}      & \multicolumn{1}{c}{\textbf{Example (Trainer)}}                                                                                                                                                                                                              & \multicolumn{1}{c}{\textbf{Example (Participant)}}                                                                                                                                                                                                                      \\ \hline
\textbf{Guideline}     & \textit{The guidelines recommend that adults get a minimum of 150 minutes, or 2.5 hours, of moderate to vigorous exercise per week.}                                                                           & \textit{I didn’t realize that I was supposed to be getting that much.}                                                                                                                                                                                                   \\ \hline
\textbf{Benefit}       & \textit{Some benefits that'll help you and everyone regardless of their BMI or age or anything like is you have decreased risk of breast and colon cancer, coronary heart disease, high blood pressure diabetes, stress, depressive symptoms, osteoporosis.} & \textit{Weight maintenance, the body image, and definitely the decrease in diabetes, stress, high blood pressure}                                                                                                                                                        \\ \hline
\textbf{Goal}          & \textit{Each week, we want you to increase your daily step count goal by 20\%.}                                                               & \textit{I would love it to be even more than that, but I think I should put my goal as to start with thirty minutes.}                                                                                                                                              \\ \hline
\textbf{Monitoring}    & \textit{How realistic is for you to get out of the house every now and then and go do ten, twelve-minute bouts, or half an hour about, whatever you need?}                                                                                                   & \textit{Sometimes I know it's hard, umm, so usually I'm off on Wednesdays and Fridays, so I can walk him three times a day.}                                                                                                                                             \\ \hline
\textbf{Support}       & \textit{Even just talking to the people around you about your goals is a fantastic first step, but it can also help to get them directly involved.}                                    & \textit{I have friends and stuff that I work with that we, we always talk about because we all have our little things and our little agendas, and always comparing notes, and, you know, just saying, "Oh, what are you doing," or, you know, "How's this?"} \\ \hline
\textbf{Self-efficacy} & \textit{If you stick to each short-term goal, I think you’ll be surprised by just how capable you really are.}                                                         & \textit{I'm pretty sure I can do that.}                                                                                                                                                                                                                                  \\ \hline
\textbf{Motivation}    & \textit{It sounds like you might be able to stay more motivated if you shake up your routine a little bit.}                                                                    & \textit{So umm, I have a couple of workouts that I can do at home if I decide I don't wanna drive out to the gym and then there's a new gym thing that's a couple of blocks down that I can try.}                                                                              \\ \hline
\textbf{Barrier}       & \textit{Has it been any easier lately to fit some physical activity into your schedule?}                                                                                                                                                                    & \textit{I mean, I said my worst thing is sometimes if I feel like I'm too busy or work is doing something over my schedule, umm, it gets a little tough.}                                                                                                                \\ \hline
\textbf{Relapse}       & \textit{What is causing you to relapse into old habits?}                                                                                                                                                                 & \textit{So I was just like, this is not definitely something I can keep up with right now.}                                                                                                                                                                              \\ \hline
\textbf{Safety}        & \textit{It’s very important to keep safety in mind while being physically active.}                                                                                    & \textit{Yeah, I try not to do that because, you know, you just make yourself a easy target.}                                                                                                                                                                             \\ \hline
\textbf{Diet}          & \textit{It’s important to choose breakfast foods that fill you up and give you long-lasting energy.}                                              & \textit{Well I've been actually the last two weeks, three weeks, or maybe it's probably when I started here, I'm with Diets-To-Go, so I'm getting that...the low carb.}                                                                                                  \\ \hline
\textbf{Weight}        & \textit{So today we want to talk about healthy weight management.}                                                                              & \textit{So according to my scale, of course, you know, there was Super Bowl Sunday on Sunday, so that probably messed everything up, I did lose some pounds.}                                                                                                            \\ \hline

\end{tabularx}
\caption{Example sentences of the strategy annotation scheme.}
\label{tab:strategy}
\end{table*}

\begin{table*}[ht]
    \small
    \begin{subtable}{.22\textwidth}
        \centering
            \begin{tabular}{lr}
            \hline
            \multicolumn{2}{l}{\textbf{Domain}} \\ \hline
            Barrier           & 2,177    \\ 
            Support           & 1,450    \\ 
            Off-task          & 1,120    \\ 
            Motivation        & 791      \\ 
            Goal              & 639      \\ 
            Safety            & 439      \\ 
            Benefit           & 346      \\ 
            Weight            & 316      \\ 
            Diet              & 185      \\ 
            Introduction      & 133      \\ 
            Guideline         & 0        \\ 
            Relapse           & 0        \\ 
            Monitoring        & 0        \\ 
            Self-efficacy     & 0        \\ \hline
            \end{tabular}
        \caption{Domain}
        \label{fig:sub1}
    \end{subtable}
    \begin{subtable}{.30\textwidth}
        \centering
            \begin{tabular}{lrr}
            \hline
            \multicolumn{2}{l}{\textbf{Strategy }} \\ \hline
                                   & \textbf{1} &  \textbf{2}     \\ 
            None                   & 4,790    &  6,901  \\ 
            Motivation             & 593      &  152   \\ 
            Support                & 542      &   39  \\ 
            Monitoring             & 374      &   236 \\ 
            Barrier                & 328      &   54 \\ 
            Safety                 & 280      &   11 \\ 
            Diet                   & 174      &   53 \\ 
            Goal                   & 169      &   71 \\ 
            Benefit                & 160      &   12 \\ 
            Self-efficacy          & 99      &   28 \\ 
            Weight                 & 72       &   25 \\ 
            Relapse                & 15       &   14 \\ 
            Guideline              & 0        &   0  \\ \hline
            \end{tabular}
        \caption{Strategy}
        \label{fig:sub2}
    \end{subtable} 
    \begin{subtable}{.22\textwidth}
        \begin{tabular}{lr}
        \hline
        \multicolumn{2}{l}{\textbf{Social Exchange}}                                   \\ \hline
        None                                                             & 5,219 \\ 
        Agree                                                            & 1,830 \\ 
        Incomplete                                                       & 350  \\ 
        \begin{tabular}[c]{@{}l@{}}Approve\\ /Encourage\end{tabular}     & 107  \\ 
        \begin{tabular}[c]{@{}l@{}}Disapprove\\ /Discourage\end{tabular} & 90   \\ \hline
        \end{tabular}
        \caption{Social exchange}
    \end{subtable}
    \begin{subtable}{.20\textwidth}
        \begin{tabular}{lr}
        \hline
        \multicolumn{2}{l}{\textbf{Task Focused}} \\ \hline
        None                    & 3,702     \\ 
        Give-GenInfo            & 2,014     \\ 
        Give-PerInfo            & 1,059     \\ 
        Ask-PerInfo             & 451      \\ 
        Give-Opinion            & 119      \\ 
        Orient                  & 95       \\ 
        Ask-GenInfo             & 56       \\ 
        Ask-Repeat              & 49       \\ 
        \begin{tabular}[c]{@{}l@{}}Check-\\ Understanding\end{tabular}     & 39       \\ 
        Ask-Opinion             & 12       \\ \hline
        \end{tabular}
        \caption{Task-Focused}
    \end{subtable}
    \caption{Annotation statistics: number of sentences annotated for the four dimensions: domain, strategy, social exchange and task-focused exchange. }
\label{tab:annotation-statistics}
\end{table*}

After the data collection, we developed an annotation scheme to categorize different conversational behaviors used by trainers and participants systematically.  The annotation scheme largely consisted of intervention-related categories and general conversational categories. Intervention-related categories included \textbf{domain} categories which were used to segment larger stretches of the conversations by topic. In addition, categories pertaining to specific \textbf{strategies} used during the intervention were included. For general conversational categories, we included \textbf{social exchange} and \textbf{task-focused} exchange categories that were borrowed from the Roter Method of Interaction Process Analysis \cite{roter1991roter}. Based on our annotation scheme, we annotated the in-person counseling sessions on a per-sentence level (sentences have been obtained using NLTK’s PunktSentenceTokenizer) across four different dimensions: domain, strategy, social exchange, and task-focused exchange. A sample dialog snippet with annotations is shown in Table \ref{tab:annotation-example}. Descriptions for the four dimensions and the included categories are as follows:

\textbf{Domain} was used to segment larger stretches (i.e., modules) of the conversations by topic. Therefore, it was coded based on the large conversational segment's overall topic, not each sentence's content. The domain categories were mainly derived from the agenda of the counseling session. 
In total, 14 domain categories were used in the study: \textit{Introduction} category covers the beginning of the conversations, \textit{Guideline} category covered conversations that refer to the physical activity guidelines for Americans, \textit{Benefit} category covered conversations addressing the health benefits of physical activity, \textit{Goal} category was related to setting short-term and long-term goals, \textit{Monitoring} category pertained to conversations on self-monitoring and adherence, \textit{Motivation} category was related to talking about staying motivated to being active, \textit{Barrier} category was about identifying and overcoming barriers to being active, \textit{Relapse} category pertained to talking about relapse and prevention, \textit{Safety} category addressed safety of physical activity, \textit{Diet} category addressed healthy diet, \textit{Weight} category denoted weight loss and maintenance, and \textit{Off-Task} category covered sustained conversations that do not fall into any of the above domain categories.

\textbf{Strategy} refers to the intention of the sentence. Categories for strategy dimension largely overlapped with categories in the domain categories except for that \textit{Introduction} category was omitted, and \textit{None} category was used instead of an Off-Task category (i.e., sentences without strategy were coded into the None category). Although the categories of the strategy and domain dimensions were very similar as they were both intervention-related, the strategies were annotated based on the specific sentence instead of the overall stretches, revealing which intervention strategies are used in the sentence. The strategies may or may not overlap with the domain. For example, the sentence ``\textit{Which benefits of physical activity are the most important to you?}'' is annotated with \textit{Benefit} for both domain and strategy, while ``\textit{How confident do you feel that you can meet your long-term goals each week?}'' belongs to the \textit{Goal} domain but has the strategy of \textit{Self-efficacy}. Considering in a few cases one sentence might belong to multiple strategies, we annotated up to two strategies (as strategy1 and strategy2) for each sentence. The order of the labeled categories was based on their relevance to the utterance. Example sentences for each strategy category are presented in Table \ref{tab:strategy}.

\textbf{Social exchange} covered personal remarks and social conversations. \textit{Greeting} and \textit{Goodbye} categories covered statements formal greetings and goodbyes. \textit{Approve/Encourage} covered positive responses such as compliments, encouragements, gratitude, and respect. \textit{Disapprove/Discourage} covered negative responses such as discouragement, criticism, and denial. \textit{Agree} category pertained to showing agreement or understanding. \textit{Incomplete} category was used only for grammatically incomplete utterances. Sentences without a social exchange were coded as ‘None’.

\textbf{Task-focused exchange} covered utterances asking for and providing information relevant to the task. \textit{Orient} category covered introductory statements about the intervention. \textit{Ask-GenInfo} and \textit{Give-GenInfo} categories covered utterances asking and providing non-personal information. On the other hand, \textit{Ask-PerInfo} and \textit{Give-PerInfo} pertained to utterances that ask and provide personal information. \textit{Ask-Opinion} and \textit{Give-Opinion} categories included utterances asking for and providing one’s subjective thoughts and feelings. Other categories included \textit{Ask-Repeat} category for sentences requesting repetition of a previous utterance and \textit{Check-Understanding} category for sentences confirming information that was just said has been understood. Sentences without task-focused content were coded as \textit{None}.

Two coders with expertise in the field annotated 17 unique in-person counseling dialogs (7,808 sentences in total). Class distributions for each dimension are shown in Table  \ref{tab:annotation-statistics}.  For domain dimension, barrier and support had the highest occurrence. For strategy, motivation is the leading one, followed by support, monitoring, and barrier. Note that a large number of sentences did not contain any strategy. As for social exchange, the amount of \textit{agree} was much higher than the others. For task-focused, most sentences were related to information-giving, especially general information (\textit{Give-GenInfo}) and personal information \textit{Give-PerInfo}.

We computed Cohen’s kappa on three double annotated in-person counseling dialogs (1,332 sentences in total) for each dimension to measure inter-rater reliability. We reach a kappa value of 0.96 for Domain, 0.76 for strategy one, 0.50 for Strategy two, 0.75 for Social Exchange, and 0.80 for Task-Focused dimensions.


\section{Strategy Classifier}

To built a dialog system capable of delivering physical activity interventions, it was first necessary to understand patterns in human-delivered intervention counseling sessions. Since the strategy dimension is intervention-related and represents each sentence's intention, in this study, we focused on examining how the strategy dimension influenced people's physical activity-related barriers and social support. Therefore, we built a BERT-based strategy classifier to leverage a large number of unannotated dialogs.

We started with the BERT-based model pre-trained on Wikipedia. We fine-tuned the model with 63,288 unannotated utterances from the physical activity counseling sessions before training on the classification task. We then trained a single-label prediction model with the 17 annotated counseling sessions (7,808 sentences in total) using leave-one-out cross-validation, where each training unit was composed of one session. 

Contextual information is crucial in dialog act predictions \cite{yu2019midas}. Hence, we considered the previous ten sentences as the dialog history. As an input to the model, we appended the history to the current sentence and used a special separate token to separate them. Table \ref{tab:annotation-statistics} shows the dataset is highly imbalanced, so we balanced the training data by randomly oversampling minority classes and undersampling majority classes. After balancing, each class had equal distribution and the size of the training set doubled. The model used 12 layers with 12 attention heads and a hidden size of 768. The fully connected layers used a dropout rate of 0.1. After training, the model reached an accuracy of 0.83 and a macro average F1 score of 0.70.

\begin{figure}[t]
\includegraphics[width=0.50\textwidth]{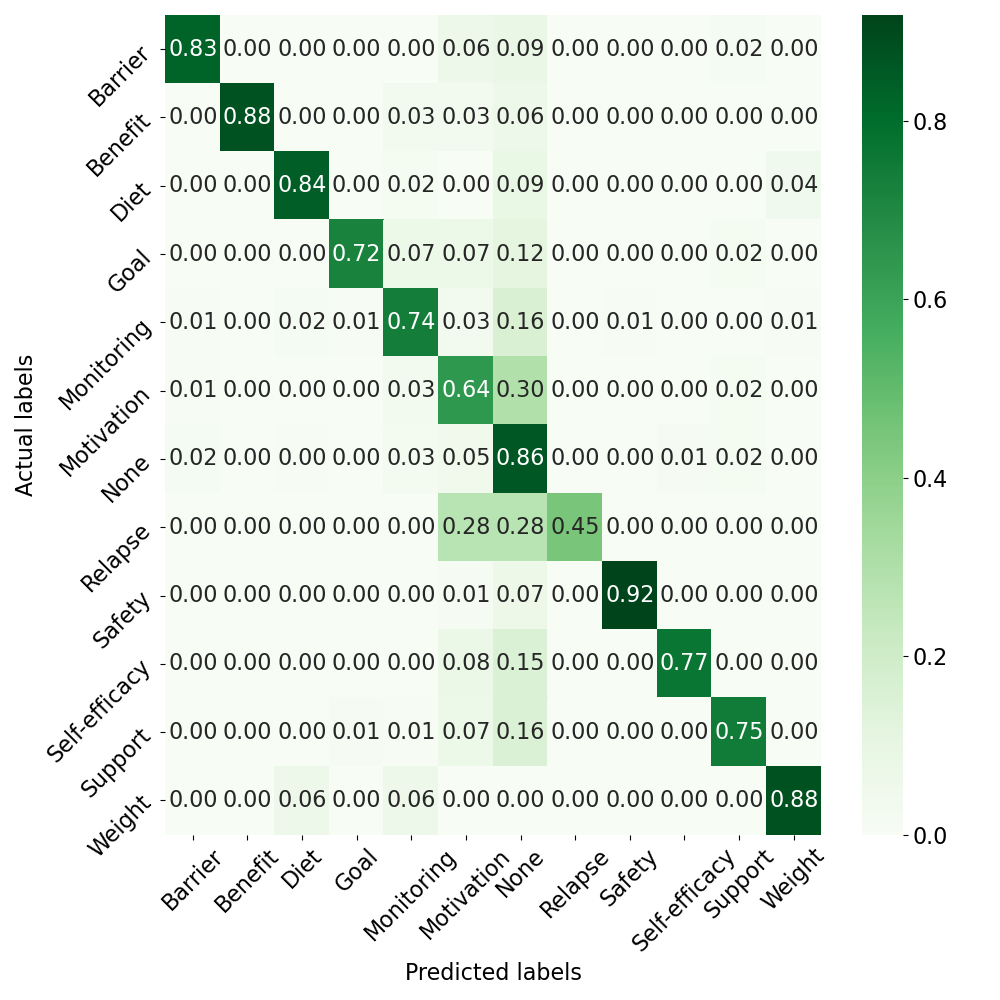}
\caption{Confusion matrix of the strategy classification.}
\label{fig:confusion_matrix}
\end{figure}

We then plotted the confusion matrix in Figure \ref{fig:confusion_matrix} to analyze the results. We found that the main error came from the misclassification of \textit{Relapse}. Relapse was sometimes classified as \textit{Motivation} mostly because people talked about recovering from relapse or staying motivated without giving up. For example, ``\textit{I was doing yoga and Pilates and needed to pick that up.}'' mentions activities that motivate the participant to recover from relapse. Another error was that \textit{Motivation} was sometimes mistaken as \textit{None} due to the diverse activities trainers mentioned to motivate the participants. 


We used the model to classify all the 88 unannotated dialogs with 63,288 sentences. The statistics are shown in Table \ref{tab:strategy_classification_results}. The distribution was similar to the annotation, where \textit{Motivation} remained the most frequent strategy, followed by \textit{Support}, \textit{Monitoring}, \textit{Safety} and \textit{Barrier}.

\begin{table}[h]
\begin{tabular}{lrrr}
\hline
\multirow{2}{*}{\textbf{Strategy}} & \multicolumn{3}{c}{\textbf{\#. Sentences}}                                                                                                                                       \\ \cline{2-4} 
                                   & \multicolumn{1}{c}{\textbf{\begin{small}\begin{tabular}[c]{@{}c@{}}Trainer \\ + Participant\end{tabular}\end{small}}} & \multicolumn{1}{c}{\textbf{\begin{small}Trainer\end{small}}} & \multicolumn{1}{c}{\textbf{\begin{small}Participant\end{small}}} \\ \hline
\textbf{None}                      & 45,079                                                                                        & 23,391                               & 21,688                                   \\
\textbf{Motivation}                & 4,012                                                                                         & 2,824                                & 1,188                                    \\
\textbf{Support}                   & 3,753                                                                                         & 2,766                                & 987                                      \\
\textbf{Monitoring}                & 3,341                                                                                         & 2,677                                & 664                                      \\
\textbf{Safety}                    & 2,049                                                                                         & 1,785                                & 264                                      \\
\textbf{Barrier}                   & 1,966                                                                                         & 1,075                                & 891                                      \\
\textbf{Diet}                      & 1,337                                                                                         & 1,136                                & 201                                      \\
\textbf{Benefit}                   & 1,158                                                                                         & 775                                  & 383                                      \\
\textbf{Goal}                      & 1,113                                                                                         & 1,003                                & 110                                      \\
\textbf{Weight}                    & 520                                                                                           & 368                                  & 152                                      \\
\textbf{Self-efficacy}             & 498                                                                                           & 200                                  & 298                                      \\
\textbf{Relapse}                   & 46                                                                                            & 29                                   & 17                                       \\ \hline
\end{tabular}
\caption{Strategy classification statistics of the classified 88 dialogs (63,288 sentences).}
\label{tab:strategy_classification_results}
\end{table}

\section{Results}
We conducted Pearson's correlation analysis to assess the relationship between the amount of barrier and support strategies and the changes in their corresponding survey scores. We also performed multiple linear regression analysis to see the strategy's effect after controlling for social-demographic factors and baseline survey scores \cite{aickin2009dealing}. 

The results are shown in Table \ref{tab:result}. We anticipated that the effect of the amount of strategies used from trainers would differ from participant, therefore we first computed each side's correlation separately. Then, the combined effect of trainers and participants was investigated. Lastly, we conducted a similar analysis to examine whether participants with heavier weight (measured at their baseline visit) used more weight strategies.


\begin{table*}[ht!]
\fontsize{8}{10}\selectfont
 \begin{adjustwidth}{}{}

\centering
\begin{tabular}{|p{2.2cm}|p{4.05cm}|p{1cm}p{1.05cm}|p{1cm}p{1.05cm}|p{1cm}p{1.05cm}|}
\hline
\multirow{2}{2cm}{\textbf{Dependent Variable}}           & \multirow{2}{2cm}{\textbf{Independent Variable}} & \multicolumn{2}{c|}{\textbf{Trainer + Participant}}                                                                                & \multicolumn{2}{c|}{\textbf{Trainer}}                                                                                               & \multicolumn{2}{c|}{\textbf{Participant}}                                                                                           \\ \cline{3-8} 
                                                       &                                                & \textbf{\begin{tabular}[c]{@{}c@{}}Pearson's\\ r\end{tabular}} & \textbf{\begin{tabular}[c]{@{}c@{}}Multple\\ Coeff.\end{tabular}} & \textbf{\begin{tabular}[c]{@{}c@{}}Pearson's \\ r\end{tabular}} & \textbf{\begin{tabular}[c]{@{}c@{}}Multple\\ Coeff.\end{tabular}} & \textbf{\begin{tabular}[c]{@{}c@{}}Pearson's \\ r\end{tabular}} & \textbf{\begin{tabular}[c]{@{}c@{}}Multple\\ Coeff.\end{tabular}} \\ \hline
\multirow{2}{2cm}{\textbf{Changes in barrier survey}}             & \textbf{\#. Barrier strategy}                           & \textbf{0.28**}                                                & \textbf{0.29**}                                                   & 0.19                                                            & 0.40                                                              & \textbf{0.27**}                                                 & \textbf{0.34*}                                                    \\ \cline{2-8} 
                                                       & Barrier survey baseline                        & -                                                              & \textbf{-0.57***}                                                 & -                                                               & \textbf{-0.58***}                                                 & -                                                               & \textbf{-0.56***}                                                 \\ \hline
\multirow{2}{2.5cm}{\textbf{Changes in support from friend survey}} & \textbf{\#. Support strategy}                           & 0.17                                                           & -0.06                                                             & 0.06                                                            & -0.11                                                             & \textbf{0.23*}                                                           & -0.08                                                             \\ \cline{2-8} 
                                                       & Support from friend survey baseline            & -                                                              & \textbf{-0.28***}                                                 & -                                                               & \textbf{-0.19*}                                                   & -                                                               & \textbf{-0.28***}                                                 \\ \hline 
\multirow{3}{2.5cm}{\textbf{Changes in support from family survey}} & \textbf{\#. Support strategy}                           & -0.16                                                          & -0.08                                                             & -0.18                                                           & -0.11                                                             & 0.10                                                            & -0.17                                                             \\ \cline{2-8} 
                                                       & Support from family survey baseline            & -                                                              & \textbf{-0.19*}                                                   & -                                                               & \textbf{-0.19*}                                                   & -                                                               & \textbf{-0.18*}                                                   \\ \cline{2-8} 
                                                       & Marriage (married)                             & -                                                              & \textbf{3.57*}                                                    & -                                                               & \textbf{3.68*}                                                    & -                                                               & 2.33                                                              \\ \cline{2-8} 
                                                       & Ethnicity (multi-race, Black and Hispanic)                                & -                                                              & \textbf{-4.84*}                                                   & -                                                               & \textbf{-4.95*}                                                   & -                                                               & \textbf{-6.59*}                                                   \\ \hline 
\multirow{1}{2.5cm}{\textbf{\#. Weight strategy}}                   & \textbf{Weight baseline}                                & 0.13                                                           & 0.012                                                             & 0.00                                                            & 0.00                                                              & \textbf{0.21*}                                                  & 0.01                                                              \\ \hline 
\end{tabular}
\caption{Results of Pearson's correlation analysis and multiple linear regression analysis.  The coefficients are calculated for different sets of dependent variables and independent variables. The ``Trainer + Participant'' column counts the corresponding amount of strategy from both speakers, where the ``Trainer'' and ``Participant'' columns counts the strategy from the trainer and participant respectively. Note that only the independent variables with significant coefficient or of main interest are shown. 
Please find full results in Table \ref{tab:results_full}.
($*: p < .05; **; p < .01; ***; p < .001$). 
}
\label{tab:result}
\end{adjustwidth}
\end{table*}

\begin{figure}[h!]
\centering
\includegraphics[width=0.45\textwidth]{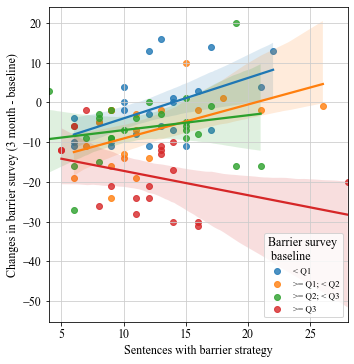}
\caption{Relationship between the amount of sentences with barrier strategies spoken by the trainer and the participants' changes in barrier survey (3 month - baseline) }
\label{fig:barrier_trainer_baseline_group}
\end{figure}

\subsection{Does using more barrier strategies decrease participants' physical activity related barriers? (RQ1)}

As shown in Table \ref{tab:result}, the number of barrier strategies used by the trainer did not have a significant effect on the changes in participants' barrier survey. However, the multiple regression analysis ($R^2=.38, F(8, 79)=5.98, p<.001$) showed that participants with a higher barrier score at the baseline visit overcame more barriers after three months ($\beta=-0.57, p<.001$), and the result remains significant after the Bonferroni multiple tests correction ($p<.001$).  This is understandable because people starting with higher barrier scores have more room for improvement, and the intervention effectively identifies and reduces their barriers. Moreover, there was significant interaction between the amount of barrier strategies used by trainers and barrier survey baseline score ($F(9, 78) = 6.46$). To investigate the interaction between them, we divided data points into four groups by the quartile values of barrier survey baseline value, where $Q_{1}$ being the lowest quartile and $Q_{3}$ the highest. As shown in Figure \ref{fig:barrier_trainer_baseline_group}, people in the group with the highest barrier baseline score overcame more barriers when the trainer used more barrier strategies, while the rest of the groups had the opposite trends. This indicates that trainers' usage of barrier strategy is beneficial for the people starting with a high barrier. Therefore, a future chatbot should discuss more barriers only to those with a very high barrier baseline. It is not recommended to do so to the rest to avoid adverse effects.

The results also showed a higher number of barrier strategies from the participants significantly predicted fewer decreases in barrier survey score ($r=0.27, p < 0.01$). The multiple regression analysis ($R^2=.38, F(8, 79)=6.16, p<.001$) showed similar results ($\beta=0.34, p=0.032$). This was interesting since the more the participants talked about their barriers, they were less likely to overcome their barriers in the end. This could mean that talking about barriers may not necessarily help them overcome them. Rather, turning the conversation to more future-directed, action-based suggestions may be more beneficial. Thus, for future chatbot development, if a participant tends to talk too much about barriers, the bot should stop discussing barriers to avoid negative effects. We also found that the participants with a higher barrier score at baseline visit overcame more barriers after three months ($\beta=-0.56, p<.001$ adjusted with Bonferroni correction). As discussed above, this may be due to the fact that they had more room for improvement. However, there was no significant interaction between the amount of barrier strategy and barrier survey baseline score ($F(9, 78)=6.18, p=n.s.$). The effect of the barrier strategy from the combination of both trainer and participant showed similar results to the participants only.

\subsection{Does using more support strategies increase participants' physical activity-related social support? (RQ2)}
To evaluate the participant's social support changes, we surveyed their support from friends and family separately. As presented in Table \ref{tab:result}, the changes in support from friends were positively correlated to the amount of support strategy from the participants ($r=0.23, p <0.5$). This means that the more the participants talked about social support, the more they gained social support from friends at the end. This suggests that a future chatbot should encourage participants to talk more about social support to achieve better outcomes. However, the effect was not significant accounting for other factors in the multiple regression model. 

The changes in support from family were not significantly correlated to the amount of support strategy regardless of the speaker. However, the analysis of overall utterances (trainer + participant) showed that women who were married gained more social support from family ($\beta=3.57, p <.05$). This suggests that a future chatbot should discuss social support from family targeting this specific demographic (i.e., married women) to gain effective outcomes. The result also showed that people belonging to multi-race, black, and Hispanic ethnicities gained less support from family ($\beta=-4.84, p <.05$). There was no interaction effect found between ethnicity and the amount of support strategy. 

Overall, participants who had lower support from friend at baseline gained more support at the end ($\beta=-0.28, p<.001$ (trainer+ participant), $\beta=-0.19$, $p<.05$ (trainer), $\beta=-0.28, p<.001$ (participant)). The results of support from family showed a similar trend ($\beta=-0.19, p<.05$ (trainer + participant), $\beta=-0.19$, $p<.05$ (trainer), $\beta=-0.18, p<.05$ (participant), while the correlation were not as high as the ones from support from friend. The increase in family support was not as high as from friends might be because people cannot change their family members, but there are more friends available to seek help.
The intervention was beneficial for participants who lacked social support to gain support from friends and family. This suggests that a future chatbot should discuss more about social support with participants who lack social support the most, especially those who lack support from friends. There was no significant interaction between the amount of barrier strategy and barrier survey baseline score.

\subsection{Do participants with heavier weight use more weight strategies? (RQ3)}
Table \ref{tab:result} demonstrates that the higher the participant's baseline weight, the more the weight strategy was used by participants ($r=0.21, p =.05$). This could be because participants with heavier weight might have had more concerns about their weight management. Thus, a future chatbot could provide more weight-related strategies towards participants with heavier weight and see if this positively affects the physical activity outcomes. Unfortunately, this effect was not significant after the adjustment in the multiple regression analysis. 
\section{Conclusions and Future Work}

In this work, we presented the foundation work on building an automatic physical activity intervention chatbot. A human-human physical activity intervention dialog dataset was created from a real intervention setting. We also designed a set of comprehensive annotation schemes and annotated the dataset at the sentence level. A strategy classifier with context embedding was shown to achieves good results on intervention strategy detection. 

The analyses showed that the amount of barrier and support strategies used in the intervention were correlated with the changes in the corresponding score, and the effects differed based on participants' baseline score and socio-demographic. We also found that people with a heavier weight at the beginning tend to talk more about weight.
Given the analysis result, we provided suggestions on designing a behavior-change intervention chatbot that could adapt to different individuals to yield better outcomes. 

This project lays the ground for the next step, which is to build a physical activity intervention chatbot that can effectively choose appropriate strategies based on user profiles and survey baseline result information to increase the intervention's effectiveness. In addition, although the main focus of this study was to investigate the association between intervention strategies and physical activity outcomes, social exchange and task-focused categories would also provide useful insights for identifying more effective conversational patterns in future studies. For example, social-exchange categories provide information on patients' acceptance towards strategies used by healthcare providers. Task-focused categories inform the exchange of information and opinions. By combining social exchange and task-focused categories with strategy information, we will be able to provide richer content and context to our interpretation of the conversation. Since the findings in our study are exploratory, we will also confirm the multiple hypotheses in the following study as pre-hoc hypotheses. 


\section*{Acknowledgements}
This project was supported by grant R01HL104147 from the National Heart, Lung, and Blood Institute and by the American Heart Association,  grant K24NR015812 from the National Institute of Nursing Research, and grant (RAP Team Science Award) from the University of California, San Francisco. The study sponsors had no role in the study design; collection, analysis, or interpretation of data; writing of the report; or decision to submit the report for publication. We also thank Ms. Kiley Charbonneau for her assistance with data management and annotations.

\bibliographystyle{acl_natbib}
\bibliography{reference}

\appendix
\clearpage
\newpage 
\appendix
\section{Appendix}

\subsection{Participant Eligibility Criteria}
\label{sec:eligibility}
Eligibility criteria for inclusion in the study were: female sex, age from 25 to 65 years, body mass index (BMI; calculated as weight in kilograms divided by height in meters squared) of 18.5 to 43.0, physically inactive at work and/or during leisure time based on the Stanford Brief Activity Survey \cite{taylor2006validation}, intent to be physically active, access to a home telephone or mobile phone, ability to speak and read English, no medical conditions or physical problems that required special attention in an exercise program, no current participation in other lifestyle modification programs, and no mild cognitive impairment as determined by the Mini-Cog test \cite{borson2000mini}.

\begin{table*}[ht!]
\fontsize{8}{10}\selectfont
\centering
\begin{tabular}{|p{2cm}|p{3.6cm}|cc|cc|cc|}
\hline
\multirow{2}{2cm}{\textbf{Dependent Variable}}           & \multirow{2}{2cm}{\textbf{Independent Variable}} & \multicolumn{2}{c|}{\textbf{Trainer + Participant}}                                                                                & \multicolumn{2}{c|}{\textbf{Trainer}}                                                                                               & \multicolumn{2}{c|}{\textbf{Participant}}                                                                                           \\ \cline{3-8} 
                                                      &                                                & \textbf{\begin{tabular}[c]{@{}c@{}}Pearson's\\ r\end{tabular}} & \textbf{\begin{tabular}[c]{@{}c@{}}Multple\\ Coeff.\end{tabular}} & \textbf{\begin{tabular}[c]{@{}c@{}}Pearson's \\ r\end{tabular}} & \textbf{\begin{tabular}[c]{@{}c@{}}Multple\\ Coeff.\end{tabular}} & \textbf{\begin{tabular}[c]{@{}c@{}}Pearson's \\ r\end{tabular}} & \textbf{\begin{tabular}[c]{@{}c@{}}Multple\\ Coeff.\end{tabular}} \\ \hline
\multirow{8}{2cm}{\textbf{Changes in barrier survey}}             & \textbf{\#. Barrier strategy}                           & \textbf{0.28**}                                                & \textbf{0.29**}                                                   & 0.19                                                            & 0.40                                                              & \textbf{0.27**}                                                 & \textbf{0.34*}                                                    \\ \cline{2-8} 
                                                      & Barrier survey baseline                        & -                                                              & \textbf{-0.57***}                                                 & -                                                               & \textbf{-0.58***}                                                 & -                                                               & \textbf{-0.56***}                                                 \\ \cline{2-8} 
                                                      & Age                                            & -                                                              & -0.17                                                             & -                                                               & -0.17                                                             & -                                                               & -0.15                                                             \\ \cline{2-8} 
                                                      & Education (college/graduate)                   & -                                                              & 1.33                                                              & -                                                               & 1.63                                                              & -                                                               & 1.55                                                              \\ \cline{2-8} 
                                                      & Ethnicity (AP)                                 & -                                                              & 1.00                                                              & -                                                               & 0.66                                                              & -                                                               & 1.10                                                              \\ \cline{2-8} 
                                                      & Ethnicity (MBH)                                & -                                                              & -4.31                                                             & -                                                               & -3.98                                                             & -                                                               & -3.80                                                             \\ \cline{2-8} 
                                                      & Marriage (married)                             & -                                                              & 0.26                                                              & -                                                               & -0.13                                                             & -                                                               & 0.08                                                              \\ \cline{2-8} 
                                                      & Employment (employed)                          & -                                                              & 2.02                                                              & -                                                               & 2.49                                                              & -                                                               & 2.68                                                              \\ \hline
\multirow{8}{2cm}{\textbf{Changes in support from friend survey}} & \textbf{\#. Support strategy}                           & 0.17                                                           & -0.06                                                             & 0.06                                                            & -0.11                                                             & \textbf{0.23*}                                                           & -0.08                                                             \\ \cline{2-8} 
                                                      & Support from friend survey baseline            & -                                                              & \textbf{-0.28***}                                                 & -                                                               & \textbf{-0.19*}                                                   & -                                                               & \textbf{-0.28***}                                                 \\ \cline{2-8} 
                                                      & Age                                            & -                                                              & -0.02                                                             & -                                                               & 0.01                                                              & -                                                               & -0.03                                                             \\ \cline{2-8} 
                                                      & Education (college/graduate)                   & -                                                              & -1.05                                                             & -                                                               & -2.85                                                             & -                                                               & -1.09                                                             \\ \cline{2-8} 
                                                      & Ethnicity (AP)                                 & -                                                              & -0.19                                                             & -                                                               & -2.09                                                             & -                                                               & -0.21                                                             \\ \cline{2-8} 
                                                      & Ethnicity (MBH)                                & -                                                              & -1.27                                                             & -                                                               & -4.95                                                             & -                                                               & -1.38                                                             \\ \cline{2-8} 
                                                      & Marriage (married)                             & -                                                              & -1.05                                                             & -                                                               & 3.68                                                              & -                                                               & -1.02                                                             \\ \cline{2-8} 
                                                      & Employment (employed)                          & -                                                              & 1.40                                                              & -                                                               & 2.33                                                              & -                                                               & 1.25                                                              \\ \hline
\multirow{8}{2cm}{\textbf{Changes in support from family survey}} & \textbf{\#. Support strategy}                           & -0.16                                                          & -0.08                                                             & -0.18                                                           & -0.11                                                             & 0.10                                                            & -0.17                                                             \\ \cline{2-8} 
                                                      & Support from family survey baseline            & -                                                              & \textbf{-0.19*}                                                   & -                                                               & \textbf{-0.19*}                                                   & -                                                               & \textbf{-0.18*}                                                   \\ \cline{2-8} 
                                                      & Marriage (married)                             & -                                                              & \textbf{3.57*}                                                    & -                                                               & \textbf{3.68*}                                                    & -                                                               & 2.33                                                              \\ \cline{2-8} 
                                                      & Ethnicity (AP)                                 & -                                                              & -2.09                                                             & -                                                               & -2.09                                                             & -                                                               & -1.87                                                             \\ \cline{2-8} 
                                                      & Ethnicity (MBH)                                & -                                                              & \textbf{-4.84*}                                                   & -                                                               & \textbf{-4.95*}                                                   & -                                                               & \textbf{-6.59*}                                                   \\ \cline{2-8} 
                                                      & Age                                            & -                                                              & 0.01                                                              & -                                                               & 0.01                                                              & -                                                               & 0.00                                                              \\ \cline{2-8} 
                                                      & Education (college/graduate)                   & -                                                              & -2.85                                                             & -                                                               & -2.85                                                             & -                                                               & -3.64                                                             \\ \cline{2-8} 
                                                      & Employment (employed)                          & -                                                              & 2.33                                                              & -                                                               & 2.33                                                              & -                                                               & 3.40                                                              \\ \hline
\multirow{7}{2cm}{\textbf{\#. Weight strategy}}                   & \textbf{Weight baseline}                                & 0.13                                                           & 0.012                                                             & 0.00                                                            & 0.00                                                              & \textbf{0.21*}                                                  & 0.01                                                              \\ \cline{2-8} 
                                                      & Age                                            & -                                                              & 0.00                                                              & -                                                               & -0.01                                                             & -                                                               & 0.01                                                              \\ \cline{2-8} 
                                                      & Education (college/graduate)                   & -                                                              & 0.08                                                              & -                                                               & 0.02                                                              & -                                                               & 0.07                                                              \\ \cline{2-8} 
                                                      & Ethnicity (AP)                                 & -                                                              & -0.39                                                             & -                                                               & 0.41                                                              & -                                                               & -0.80                                                             \\ \cline{2-8} 
                                                      & Ethnicity (MBH)                                & -                                                              & 1.82                                                              & -                                                               & 0.50                                                              & -                                                               & 1.32                                                              \\ \cline{2-8} 
                                                      & Marriage (married)                             & -                                                              & 0.30                                                              & -                                                               & -0.28                                                             & -                                                               & 0.58                                                              \\ \cline{2-8} 
                                                      & Employment (employed)                          & -                                                              & -0.59                                                             & -                                                               & -0.12                                                             & -                                                               & -0.47                                                             \\ \hline
\end{tabular}
\caption{Results of Pearson's correlation analysis and multiple linear regression analysis.  The coefficients are calculated for different sets of dependent variables and independent variables. The ``Trainer + Participant'' column counts the corresponding amount of strategy from both speakers, where the ``Trainer'' and ``Participant'' columns counts the strategy from the trainer and participant respectively. 
($*$: p $<$ 0.05, $**$: p $<$ 0.01 and $***$: p $<$ 0.001)
\\Ethnicity (AP): Asian and Pacific islander; Ethnicity (MBH): multi-race, Black and Hispanic. 
}
\label{tab:results_full}
\end{table*}

\subsection{Multiple Linear Regression Analysis Results}
Please find the full multiple linear regression analysis results in Table \ref{tab:results_full}.

\subsection{Simulated Dialog Statistics}
\label{sec:simulated-dialog}
The annotation distributions of the simulated dialogs are demonstrated in Table \ref{tab:annotation-statistics-simulated}.

\begin{table*}[ht]
    \small
    \begin{subtable}{.22\textwidth}
        \centering
            \begin{tabular}{lr}
            \hline
            \multicolumn{2}{l}{\textbf{Domain}} \\ \hline
            Barrier           & 31    \\ 
            Support           & 43    \\ 
            Off-task          & 0    \\ 
            Motivation        & 37      \\ 
            Goal              & 63      \\ 
            Safety            & 42      \\ 
            Benefit           & 29      \\ 
            Weight            & 61      \\ 
            Diet              & 43      \\ 
            Introduction      & 148      \\ 
            Guideline         & 111        \\ 
            Relapse           & 63        \\ 
            Monitoring        & 60        \\ 
            Self-efficacy     & 41        \\ \hline
            \end{tabular}
        \caption{Domain}
        \label{fig:sub1}
    \end{subtable}
    \begin{subtable}{.30\textwidth}
        \centering
            \begin{tabular}{lr}
            \hline
            \multicolumn{2}{l}{\textbf{Strategy }} \\ \hline
            None                   & 301        \\ 
            Motivation             & 68          \\ 
            Support                & 32           \\ 
            Monitoring             & 108          \\ 
            Barrier                & 74          \\ 
            Safety                 & 17          \\ 
            Diet                   & 21          \\ 
            Goal                   & 51          \\ 
            Benefit                & 22          \\ 
            Self-efficacy          & 29          \\ 
            Weight                 & 23           \\ 
            Relapse                & 9           \\ 
            Guideline              & 17          \\ \hline
            \end{tabular}
        \caption{Strategy}
        \label{fig:sub1}
    \end{subtable} 
    \begin{subtable}{.22\textwidth}
        \begin{tabular}{lr}
        \hline
        \multicolumn{2}{l}{\textbf{Social Exchange}}                                   \\ \hline
        None                                                             & 4639 \\ 
        Agree                                                            & 151 \\ 
        Incomplete                                                       & 0  \\ 
        \begin{tabular}[c]{@{}l@{}}Approve\\ /Encourage\end{tabular}     & 87  \\ 
        \begin{tabular}[c]{@{}l@{}}Disapprove\\ /Discourage\end{tabular} & 20   \\
        Greeting                                                         & 50  \\
        Goodbye                                                          & 1  \\
        \hline
        \end{tabular}
        \caption{Social exchange}
    \end{subtable}
    \begin{subtable}{.20\textwidth}
        \begin{tabular}{lr}
        \hline
        \multicolumn{2}{l}{\textbf{Task Focused}} \\ \hline
        None                    & 264     \\ 
        Give-GenInfo            & 163     \\ 
        Give-PerInfo            & 168     \\ 
        Ask-PerInfo             & 45      \\ 
        Give-Opinion            & 60      \\ 
        Orient                  & 37       \\ 
        Ask-GenInfo             & 12       \\ 
        Ask-Repeat              & 7       \\ 
        \begin{tabular}[c]{@{}l@{}}Check-\\ Understanding\end{tabular}     & 1       \\ 
        Ask-Opinion             & 15       \\ \hline
        \end{tabular}
        \caption{Task Focused}
    \end{subtable}
    \caption{Annotation statistics of the simulated dialog: number of sentences annotated for the four dimensions: domain, strategy, social exchange and task focused. }
\label{tab:annotation-statistics-simulated}
\end{table*}





\end{document}